\documentclass[final]{cvpr}

\usepackage{times}
\usepackage{epsfig}
\usepackage{graphicx}
\usepackage{amsmath}
\usepackage{amssymb}
\usepackage{multirow}
\usepackage{comment}
\usepackage{siunitx}
\usepackage{cuted}
\usepackage{caption}
\usepackage{float}
\usepackage{textcomp}
\usepackage{xcolor}
\newcommand{\RNum}[1]{\uppercase\expandafter{\romannumeral #1\relax}}
\definecolor{blue}{HTML}{0000FF}

\DeclareMathOperator{\EX}{\mathbb{E}}

\usepackage[pagebackref=true,breaklinks=true,colorlinks,bookmarks=false]{hyperref}
\usepackage{cleveref}

\def\cvprPaperID{3915} 
\def\confYear{CVPR 2021}

\begin{document}

\title{Coming Down to Earth: Satellite-to-Street View Synthesis for Geo-Localization}

\author{Aysim Toker\qquad Qunjie Zhou\qquad Maxim Maximov\qquad Laura Leal-Taixé\\
Technical University of Munich\\
{\tt\small \{aysim.toker, qunjie.zhou, maxim.maximov, leal.taixe\}@tum.de}
}

\twocolumn[{%
\renewcommand\twocolumn[1][]{#1}%
\vspace{-3.5em}
\maketitle
\thispagestyle{empty}
\begin{center}
    \centering
    \includegraphics[width=1.\textwidth, trim={0 0px 0px 0}, clip]{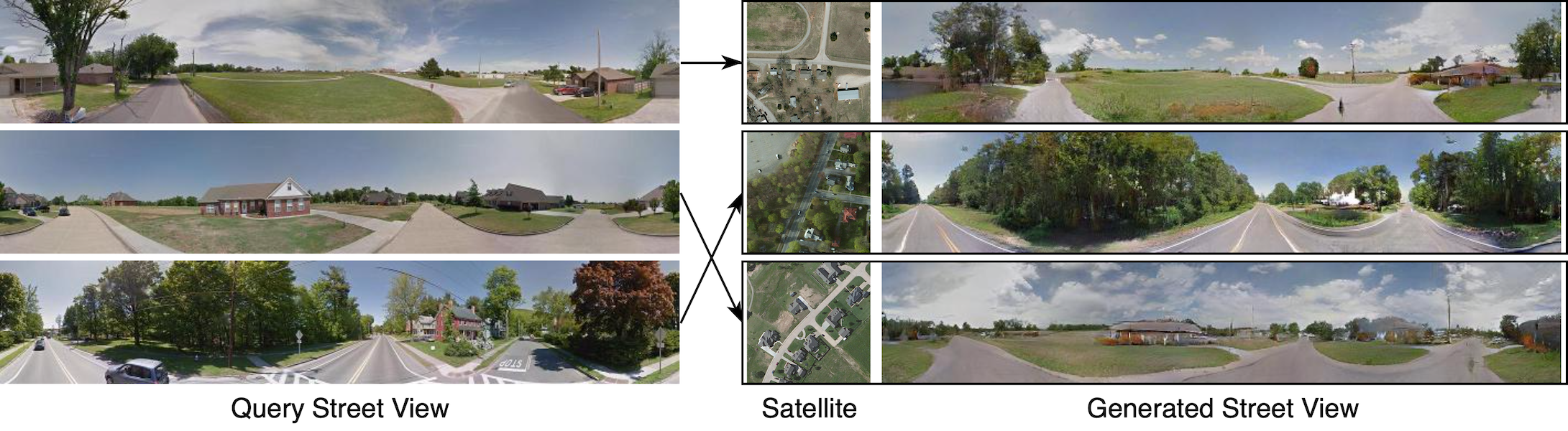}
    \captionof{figure}{
    For a collection of satellite and street images, our method synthesizes the street view for each satellite input (right). It also simultaneously determines the geographic location of a query street image by matching it with the closest satellite image in the database (left$\rightarrow$right). This is done in one single architecture which allows for end-to-end training.
    }
    \label{fig:teaser}
\end{center}%
}]
\begin{abstract}
The goal of cross-view image based geo-localization is to determine the location of a given street view image by matching it against a collection of geo-tagged satellite images. This task is notoriously challenging due to the drastic viewpoint and appearance differences between the two domains. 
We show that we can address this discrepancy explicitly by learning to synthesize realistic street views from satellite inputs. Following this observation, we propose a novel multi-task architecture in which image synthesis and retrieval are considered jointly. 
The rationale behind this is that we can bias our network to learn latent feature representations that are useful for retrieval if we utilize them to generate images across the two input domains. 
To the best of our knowledge, ours is the first approach that creates realistic street views from satellite images and localizes the corresponding query street-view simultaneously in an end-to-end manner. 
In our experiments, we obtain state-of-the-art performance on the CVUSA and CVACT benchmarks. Finally, we show compelling qualitative results for satellite-to-street view synthesis.
\end{abstract}
\section{Introduction} 
Estimating the geographic location of 
an image is a fundamental problem in computer vision with applications in autonomous driving, robotics, and augmented reality.
Originally, the problem was cast as an image retrieval task~\cite{schindler2007CVPR, hays2008CVPR, zamir2010ECCV, chen2011CVPR, arandjelovic2014ACCV, torii2015CVPR, zamir2014image, sattler2016CVPR}, where the goal is to determine the geographic location of a query street view image by comparing it against a database of GPS-tagged street images.
The main limitation of this approach is that, even though there are large databases available for this type of imagery, the coverage varies a lot between different regions of the world, and it is generally sparse in rural areas.

Satellite imagery, on the other hand, is broadly available for most parts of the world with services like Google maps. This encouraged researchers to focus on cross-view image-based geo-localization~\cite{workman2015ICCV,lin2015CVPR, vo2016ECCV,liu2019CVPR,shi2019NeurIPS,cai2019ICCV,shi2020AAAI,shi2020CVPR} as a more general and inclusive alternative. 
The overall idea is to predict the latitude and longitude of a street-level image by matching it against a GPS-tagged satellite database.
Even though this approach helps to cover vast parts of the world, the significant domain gap between a pair of street view and top-view satellite images, shown in~\Cref{fig:teaser}, makes cross-view image based geo-localization extremely challenging. 
For instance, the appearance of the two images can vary significantly as they are typically taken at different times and with different cameras, leading to illumination changes. 
The biggest challenge, however, comes from the dramatically different viewpoints of street and satellite images -- even for human eyes, it is far from obvious that two images show the same location.
Satellite images cover a broader area in comparison to the ego-centric viewpoint of the street images. On the other hand, there are a lot of additional features in street view images, like facades, that are not visible in the top-view satellite images which would otherwise be extremely useful for precise location retrieval.

In order to alleviate the difficulty of learning cross-view features, ~\cite{shi2019NeurIPS, shi2020CVPR} use a simple polar coordinate transformation as a preprocessing step for image retrieval. Intuitively, this mimics the real viewpoint transformation from the over-head view to the ground-view. Nevertheless, there is still a significant appearance gap between polar-transformed and real street images. The two views do not overlap perfectly, which limits the retrieval performance.
In the last few years, Generative Adversarial Networks (GANs)~\cite{goodfellow2014NeurIPS} have proven to be a powerful tool for generating realistic looking images. Recent works~\cite{regmi2018CVPR, regmi2019CVIU, zhu20183DV} applied them for cross-view image synthesis between aerial and ground-level images but they do not evaluate their effectiveness for the geo-localization task.~\cite{regmi2019ICCV} is the first to use pre-trained synthesized images~\cite{regmi2018CVPR} to train a retrieval network for geo-localization. However, this is done in two stages and therefore does not allow for end-to-end training. They obtained less accurate retrieval results than methods based on polar transformations~\cite{shi2019NeurIPS, shi2020CVPR}.
%
%
This suggests that, while GANs create images that \emph{look more realistic}, 
polar-transformation is more suitable to map the \emph{content} of the images across the two domains. 
In this work, our goal is to address the drastic viewpoint difference of the two domains by synthesizing realistic-looking and content-preserving street images from their satellite counterparts for geo-localization.
%
To that end, we integrate a cross-view synthesis module and a geo-localization branch in a single architecture.
The main insight here is that these two network components mutually reinforce each other: %
Learning to generate street images from satellite inputs naturally helps the image retrieval branch, since our network learns to extract local features that are useful across the two input domains.
Vice versa, the retrieval branch incentivizes our network to create realistic street views that replicate the content of a given satellite image. Additionally, our network uses polar transformed satellite images as a starting point (\ie, as an input to the GAN). This makes the image generation easier, since the spatial layout of the polar transformed image and the street view is approximately the same.

\paragraph{Contribution}
We propose a novel geo-localization method that is trained jointly for the multi-task setup of both synthesizing ground images from satellite images and retrieving cross-view image matches. We devise a single network for both of these tasks which can be trained in an end-to-end manner. 
Our method shows strong empirical results, both in terms of the retrieval accuracy and synthesis quality. For geo-localization, we obtain state-of-the-art performance on standard large-scale cross-view retrieval benchmarks. Moreover, our pipeline generates highly realistic street views that strongly resemble real, panoramic street images. Remarkably, our method outperforms existing cross-view synthesis approaches that use semantic labels as supervision during training.
\section{Related work}
The main challenge of cross-view image based geo-localization is the drastic appearance and viewpoint differences between satellite and street view image pairs.
For our discussion of existing work, we distinguish between methods that directly extract viewpoint invariant features on the input images and methods that apply an explicit viewpoint transformation to the inputs.
\paragraph{Domain-invariant features} 
A central question in cross-view image based geo-localization is how to extract features that are invariant to the appearance gap between satellite and street view images.
Early works~\cite{lin2013CVPR, mousavian2016semantic, castaldo2015CVPRW, bansal2011geo} built hand-crafted pipelines where extracted features have an explicit semantic interpretation, like detecting buildings.

Following the success of deep learning for several
computer vision tasks, new approaches successfully applied deep convolutional neural networks (CNN)
to learn feature representations for cross-view image based geo-localization.
The first approaches in this line of work were based on the AlexNet~\cite{krizhevsky2012NeurIPS} model pretrained on ImageNet~\cite{deng2009CVPR} and the Places~\cite{zhou2014NeurIPS} dataset. Originally, the pretrained weights were used directly to match image pairs without any additional training~\cite{workman2015CVPRW}. Further improvements were proposed in~\cite{workman2015ICCV}, which refined the features of the satellite images to make them more coherent with the pretrained descriptors on the street level.~\cite{lin2015CVPR} utilized a siamese network to learn features for both street view and \ang{45} aerial images with a contrastive loss.
In subsequent work,~\cite{vo2016ECCV} explored several CNN architectures and concluded that a triplet CNN trained with a soft-margin triplet loss is most suitable to extract deep features from cross-view image pairs. 

Most of these approaches used a standard fully connected layer to combine local features into a global feature representation. In contrast to that,~\cite{hu2018CVPR} embedded a learnable NETVLAD~\cite{arandjelovic2016CVPR} layer to aggregate local CNN features.~\cite{liu2019CVPR} showed that orientation information, in the form of hand-crafted UV maps, helps to convey the approximate viewpoint difference to the network during training. Recently,~\cite{cai2019ICCV} applied both spatial and channel-wise attention to the feature maps and trained them with a hard exemplar reweighting triplet loss. 

Despite the abundant progress in improving the architecture and losses for learning cross-view feature representations, trying to overcome the domain gap purely via feature learning remains a challenging open problem.
\paragraph{Viewpoint transformation}
Instead of focusing merely on the feature representation, recent approaches transformed the input images to explicitly address the viewpoint discrepancy between satellite and street view images.

The first among these approaches~\cite{zhai2017CVPR} synthesized street-level information from top-view satellite inputs. In particular, they learned to map semantic labels from the satellite view to the ground view and use them to create street view information.
Similarly,~\cite{regmi2018CVPR} applied a conditional GAN that creates ground images from the satellite view and vice versa.
~\cite{lu2020CVPR} proposed to generate street views from satellite images by utilizing depth maps and semantic labels.
Even though all of these approaches generate novel viewpoints between satellite and street observations, they do not explicitly apply it to the geo-localization problem.

A key formalism towards closing the domain gap between top-view and ground-level images was proposed by ~\cite{shi2019NeurIPS}. The main insight here is that the viewpoint transformation can be approximated with a simple change of coordinates -- in particular, a polar coordinate transformation. 
This approximately preserves the content of the satellite images, but the resulting images are far from realistic street views.
%
Using the transformed images for retrieval,~\cite{shi2019NeurIPS} further proposed to learn spatial attention maps to aggregate the local CNN features into global image descriptors.
On top of the polar transformation,~\cite{shi2020CVPR} trained a siamese network with a dynamic matching module to learn discriminative feature representations along the horizontal direction. To that end, features from the ground view are shifted such that they correlate with the polar transformed images. 

More realistic street-view images can be generated with GANs~\cite{regmi2018CVPR} which was later used as an additional input to train a retrieval network~\cite{regmi2019ICCV}.
While the obtained images look realistic, this approach lacks a strong incentive for the image generation to preserve the content of the input images, which negatively impacts the retrieval performance.
Consequently, most existing cross-view synthesis approaches require semantic maps for a sufficient preservation of content~\cite{regmi2018CVPR,zhai2017CVPR,lu2020CVPR}.

In this paper, we take a different approach. We show that we do not need semantic maps to obtain realistic looking and content-preserving street-from-satellite images. 
We observe that existing works treat the tasks of image synthesis and retrieval separately, even though the synergies are clear.
We show that our proposed multi-task training of image synthesis and retrieval in an end-to-end manner leads to state-of-the-art results, both in terms of geo-localization and cross-view image generation. 

%


\section{Method}\label{sec:method}
\begin{figure*}
\begin{center}
    \includegraphics[width=1.07\linewidth]{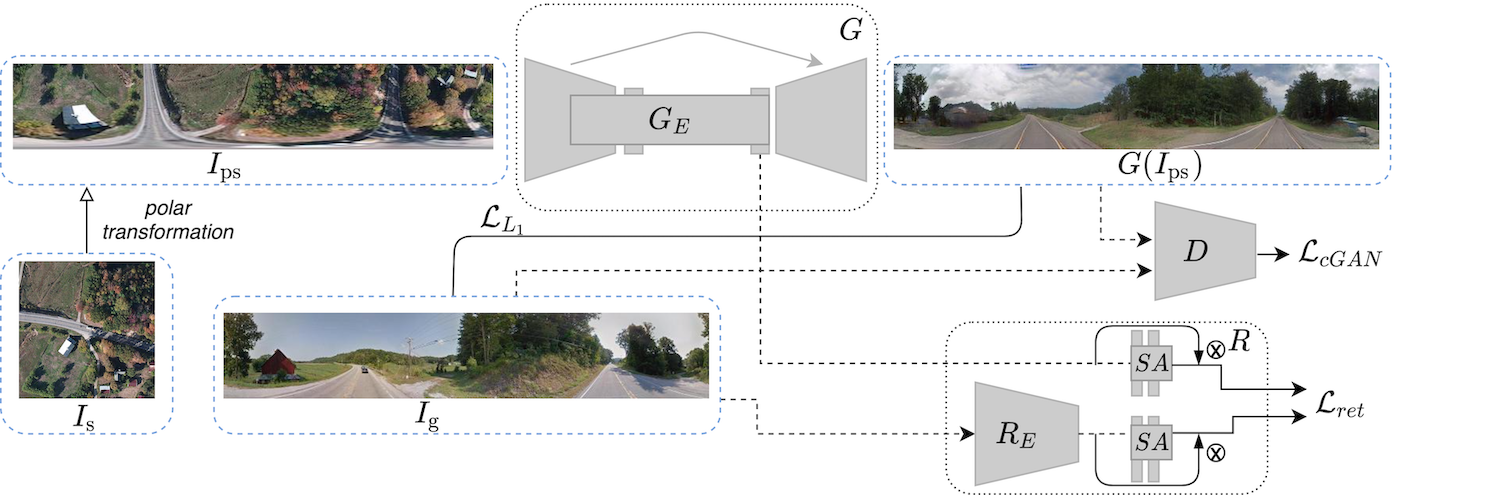}
\end{center}
   \caption{An overview of our network. We convert the pixel coordinates of the top-view satellite image $I_{\mathrm{s}}$ to $I_{\mathrm{ps}}$. Then our generative network $G$ synthesizes the street image
   $G(I_{\mathrm{ps}})$. In the same forward pass, the network feeds the projected satellite features $G_{E}(I_{\mathrm{ps}})$ and the corresponding ground image $I_{\mathrm{g}}$ to the retrieval branch. Network $R_{E}$ extracts the local features from the real street view analogous to $G_{E}$. $SA$ is a spatial-aware attention module that aggregates the extracted local features into global image descriptors. $\mathcal{L}_{cGAN}$, $\mathcal{L}_{L_{1}}$, $\mathcal{L}_{ret}$ are the loss functions that we used for learning, see~\Cref{subsec:training}.
   }
\label{fig:network_overview}
\end{figure*}
In this section, we describe our proposed multi-task approach to geo-localization, see~\Cref{fig:network_overview} for an overview.
The main idea is to jointly address the cross-view image retrieval and satellite-to-street view synthesis in a single framework. 
Specifically, we project a given pair of satellite and street images into their latent feature space and use those features simultaneously for both tasks. On one hand, the retrieval branch makes sure that the \emph{content} of the generated images is true to the real scene depicted. At the same time, the image synthesis biases our model to learn features that are consistent across the two input domains which, in turn, benefits the localization.

Initially, we apply a polar transformation to the satellite inputs~\cite{shi2020CVPR, shi2019NeurIPS}, which maps their content to an approximate street view, see~\Cref{subsec:polar}.
We then synthesize a realistic street view from the polar-transformed images, see~\Cref{subsec:gan}. At the same time, the network learns to set satellite-street pairs in correspondence in the image retrieval branch, which we outline in~\Cref{subsec:retrieval}. Finally, we provide details on the learning procedure in~\Cref{subsec:training}. Also, see our supplementary material for more technical implementation details.
%
\subsection{Polar transformation}\label{subsec:polar}
 As shown in earlier work~\cite{shi2019NeurIPS,shi2020CVPR}, we can partially bridge the domain gap of our input pairs with a simple polar coordinate transformation of the top-view satellite inputs:
\begin{equation}\label{eq:polartransform}
\begin{aligned}
x_{i}^{\mathrm{s}} = \frac{W_{\mathrm{s}}}{2} + \frac{W_s}{2} \frac{y_{i}^{\mathrm{ps}}}{H_{\mathrm{ps}}} \sin\left({\frac{2\pi}{W_{\mathrm{ps}}}x_{i}^{\mathrm{ps}}}\right)\\
y_{i}^{\mathrm{s}} = \frac{H_{\mathrm{s}}}{2} - \frac{H_s}{2} \frac{y_{i}^{\mathrm{ps}}}{H_{\mathrm{ps}}}
\cos\left({\frac{2\pi}{W_{\mathrm{ps}}}x_{i}^{\mathrm{ps}}}\right)
\end{aligned}
\end{equation} 
Here, $(x_{i}^{\mathrm{s}},y_{i}^{\mathrm{s}})$ and $(x_{i}^{\mathrm{ps}},y_{i}^{\mathrm{ps}})$ are pixel coordinates of the satellite and polar transformed images, respectively. The dimensions are specified by $W_{\mathrm{s}}\times H_{\mathrm{s}}$ and $W_{\mathrm{ps}}\times H_{\mathrm{ps}}$.
In this formulation, circular lines in the top-view satellite images become horizontal lines in the ground view. Vice-versa, radial lines correspond to vertical lines in the new set of coordinates.
In particular, the north-line, which is a vertical line originating from the center of the satellite image, corresponds to the vertical line at $\frac{W_{\mathrm{ps}}}{2}$ in the transformed image. 

Overall, this transformation produces image pairs that respect the content of the scene, \ie, they have roughly the same arrangement of objects in the scene. 
However, that alone is not sufficient to completely close the domain gap between the two views: The overlap is typically not perfect and a lot of features, like, \eg, the sky as seen from the ground-view, can simply not be recovered in that manner. Consequently, in the next step, we convert the polar transformed images to street images using a generative model.
\subsection{Generator and discriminator networks}\label{subsec:gan}
Generative Adversarial Networks (GANs)~\cite{goodfellow2014NeurIPS} are nowadays broadly used for image synthesis tasks in computer vision. The main appeal of this class of architectures is that they are able to generate highly realistic images. 
This is typically done via adversarial training of two opposing networks, the generator $G$ and the discriminator $D$.
We follow the lines of recent conditional GAN method~\cite{isola2017CVPR} since our goal is to synthesize realistic street views that, at the same time, replicate the content from a reference satellite image. 

The first component of our model is the \emph{generator} $G$ which takes a polar-transformed satellite image $I_\mathrm{ps}$ as an input and translates it into a photo-realistic street panorama $G(I_\mathrm{ps})$. The polar-coordinate representation, in this context, is a highly useful preprocessing step since the general outline of the transformed image already resembles the actual street view. This takes some of the burden of bridging the satellite-street domain gap from the generator. The generated images $G(I_\mathrm{ps})$, as well as the ground-truth street views $I_\mathrm{g}$, are then fed to the \emph{discriminator} $D$ which tries to determine whether the respective images are real or fake. The feedback from this discriminator in turn incentivizes the generator to create images that are indistinguishable from real street views. 

In the remainder of this section, we briefly outline the architecture of the two network components $G$ and $D$. For further details, we refer the interested reader to our supplementary material.
\paragraph{Generator}
Our generator network $G$ is designed as a U-Net~\cite{ronneberger2015unet} architecture, which
consists of residual blocks~\cite{he2016CVPR}. 
The first few downsampling layers, together with the network bottleneck, are called the image encoder $G_{E}$. Specifically, $G_{E}$ consists of 3 residual downsampling blocks that reduce the spatial size by a factor of 4 each. On this reduced resolution, the bottleneck layers further refine the latent features with 6 residual blocks. 
In the remainder of the generator $G\setminus G_{E}$ we use 3 residual upsampling blocks to obtain a synthesized street-level image $G(I_\mathrm{ps})$ with the same resolution as the polar-transformed input image $I_\mathrm{ps}$.
Between all downsampling and upsampling blocks we use skip connections as a standard trick to improve the network's convergence. Furthermore, we use instance normalization~\cite{ulyanov2016instance} after each residual block and spectral normalization~\cite{miyato2018spectral} after each convolution layer.
\paragraph{Discriminator}
%
%
We construct the discriminator $D$ as a PatchGAN~\cite{isola2017CVPR,li2016ECCV} classifier. 
For a given $H_\mathrm{ps}\times W_\mathrm{ps}$ street-view image, the discriminator $D$ downscales the spatial size to smaller patches and classifies each patch as either real or fake. 
%
The patch-wise strategy is particularly beneficial for synthesizing street view images, which typically consist of recurring patterns of streets, trees, and buildings. Since the global coherency is secondary in this context, the classifier can place a higher emphasis on fine-scale details.
\subsection{Retrieval network}\label{subsec:retrieval}
Having defined our image synthesis module, we now describe our retrieval branch $R$. The goal is to localize a given query street image $I_\mathrm{g}$ by matching it against a database of satellite images.
$R$ consists of two parts: An encoder block $R_{E}$ for $I_\mathrm{g}$ and a spatial attention module $SA$ that converts obtained local features of street and satellite images into global descriptors. 
For $R_{E}$, we use a modified ResNet34~\cite{he2016CVPR} backbone which extracts local features for the street-view input. We do not, however, compute an analogous latent encoding of the satellite inputs $I_\mathrm{ps}$ here. 
Instead, we reuse the features from the generator encoder $G_{E}(I_\mathrm{ps})$.

This is the core idea of our \emph{multi-task} setup: By using the learned features $G_{E}(I_\mathrm{ps})$ for both the synthesis and retrieval tasks, we allow these two aspects of the learning procedure to interact and reinforce each other. The retrieval part by itself is limited to detect and identify similar objects. The learned features from the image synthesis task, on the other hand, provide an explicit notion of domain transfer, since we learn to translate images across the two domains. In turn, the retrieval network compels the generator branch to learn features that are eventually useful for image matching -- this yields realistic generated images that also faithfully depict the content of the scene. 
\paragraph{Spatial-aware feature aggregation} 
The generator and retrieval feature encoders $G_{E}$ and $R_{E}$ learn local feature representations on both the polar-transformed satellite and the street images. In order to convert these local features $F_{\mathrm{ps}}:=G_{E}(I_{\mathrm{ps}})$ into a global descriptor $\tilde{F}_{\mathrm{ps}}$, we use a spatial-aware feature aggregation~\cite{shi2019NeurIPS} layer. 
For a given set of input features, this module predicts $k$ spatial attention masks $A_1,\dots,A_k\in\mathbb{R}^{H\times W}$. These masks $A_i$ are obtained by max-pooling $F_{\mathrm{ps}}\in\mathbb{R}^{H\times W\times C}$ along the channel dimension $C$ and refining the obtained features with two consecutive full-connected layers. The global feature components $\tilde{F}_{\mathrm{ps},i}\in\mathbb{R}^{C}$ are then defined as a weighted combination of the input features and the attention masks $A_i$:
\begin{equation}
    \tilde{F}_{ps,i}:=\bigl\langle F_{ps}, A_i\bigr\rangle_F.
\end{equation}
Here, $\langle\cdot,\cdot\rangle_F$ denotes the Frobenius inner product.
Finally, we obtain a global descriptor $\tilde{F}_{\mathrm{ps}}$ by stacking $\tilde{F}_{ps,1},\dots,\tilde{F}_{ps,k}$ into one $kC$-dimensional feature vector.
\subsection{Learning}\label{subsec:training}
The goal of our method is to jointly retrieve the correct satellite match for a given query street view, as well as synthesizing the corresponding street view from the satellite image.
To that end, we devise the following loss function:
\begin{equation}\label{eq:general}
    \mathcal{L} =  \lambda_{cGAN}\mathcal{L}_{cGAN} + \lambda_{L_1}\mathcal{L}_{L_1} + \lambda_{ret}\mathcal{L}_{ret}.
\end{equation}
During training, we then update the weights of the three components of our model $G$, $D$ and $R$ in an adversarial manner:
\begin{equation}\label{eq:minmax}
    \min_{G,R}\max_D \mathcal{L}(G,R,D).
\end{equation}
In the remainder of this section, we describe in detail how the three components of our composite loss in~\Cref{eq:general} are defined.
\paragraph{Conditional GAN loss}
For the image generation task, we define a conditional GAN loss~\cite{isola2017CVPR}:
\begin{equation}\label{eq:conditional_gan}
\begin{split}
    \mathcal{L}_{cGAN}(G, D) &= \EX_{I_{\mathrm{ps}}, I_{\mathrm{g}}}\bigl[\log D(I_{\mathrm{ps}}, I_{\mathrm{g}})\bigr] + \\
     & \EX_{I_{\mathrm{ps}}}\bigl[\log (1-D(I_{\mathrm{ps}}, G(I_{\mathrm{ps}})))\bigr].
\end{split}
\end{equation}
While the discriminator $D$ tries to classify images into real (for $I_\mathrm{g}$) and fake (for $G(I_{\mathrm{ps}})$), the generator $G$ tries to minimize the loss by creating realistic images. The corresponding satellite image $I_\mathrm{ps}$ is applied as a condition to both the discriminator and the generator.
\paragraph{Reconstruction loss}
The second component in~\Cref{eq:general} is a $L_1$ reconstruction loss which minimizes the distance between the predicted $G(I_{ps})$ and the ground-truth street-level images $I_{g}$:
\begin{equation}\label{eq:l1}
\begin{split}
    \mathcal{L}_{L_1}(G)= \EX_{I_{\mathrm{g}}, I_{\mathrm{ps}}} \bigl[ \|I_{\mathrm{g}} - G(I_{\mathrm{ps}}) \|_{1}\bigr].
\end{split}
\end{equation}
While, in principle, $\mathcal{L}_{cGAN}$ suffices to obtain meaningful translations, $\mathcal{L}_{L_1}$ still helps the network to capture low-level image features and thereby steers the image synthesis to convergence.
%
\paragraph{Retrieval loss}
Finally, we use a supervised retrieval loss for the geo-localization task, which is specified as a weighted soft-margin ranking loss~\cite{hu2018CVPR}:
\begin{flalign}\label{eq:weigthed_soft_margin}
    \mathcal{L}_{ret}(G_{E}, R_{E}, SA)=& \\\EX_{I_{\mathrm{ps}}, I_{\mathrm{g}}} \EX_{\tilde{I_{\mathrm{g}}}\neq I_{\mathrm{g}}}&\bigl[\log (1+
    e^{\alpha \mathrm{d}(I_{\mathrm{g}}, I_{\mathrm{ps}})-\alpha \mathrm{d}(\tilde{I_{\mathrm{g}}}, I_{\mathrm{ps}})})\bigr]\nonumber.
\end{flalign}
Here, the distance metric between a pair of ground and satellite images $I_g$ and $I_{ps}$ is defined as the squared $L_2$ distance between the learned features of both images:
\begin{equation}\label{eq:distancemetric}
    \mathrm{d}(I_{\mathrm{g}}, I_{\mathrm{ps}}):=\|SA(R_{E}(I_{\mathrm{g}}))-SA(G_{E}(I_{\mathrm{ps}}))\|^2_2.
\end{equation}
Intuitively, $\mathcal{L}_{ret}$ aims at decreasing the distance of positive matches in the latent space and pushes negative pairs apart. 
%
%
%
%
\begin{figure*}
    \centering
    \includegraphics[width=\linewidth]{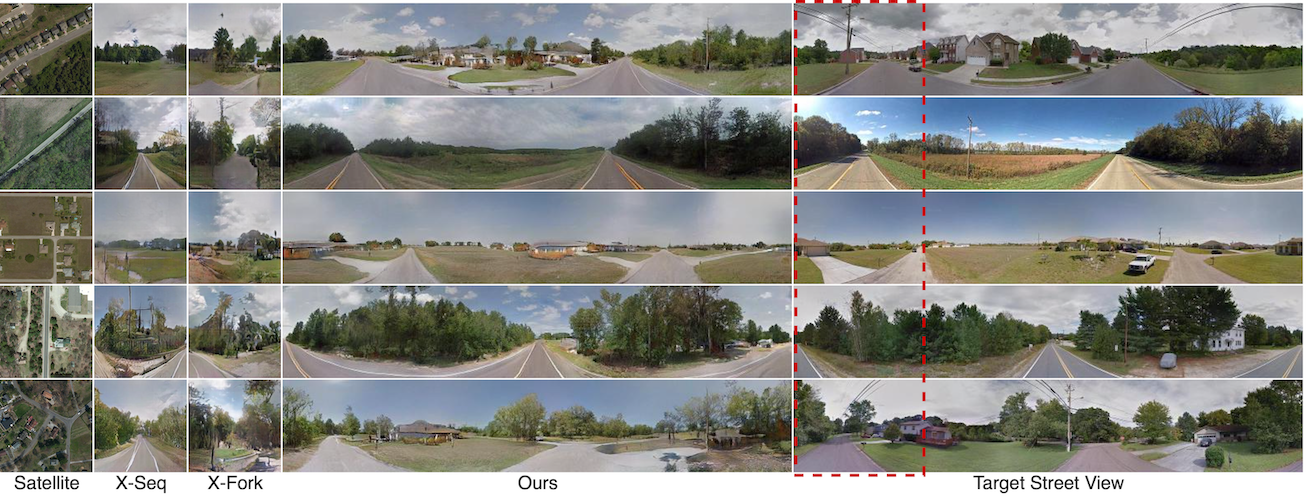}
    \caption{Qualitative comparisons for cross-view image synthesis on the CVUSA benchmark. We compare the images generated by our method with the best baselines X-Fork and X-Seq~\cite{regmi2018CVPR}. Note, that they focus on synthesizing the first quarter of the street view (which is equivalent to the red, dashed boxes on the target street-view), our method is able to create coherent full street view panoramas.}
    \label{fig:qualitative_cvusa_comparison}
\end{figure*}
\section{Experiments}
\begin{table*}[]
\begin{center}
\scalebox{0.95}{
\begin{tabular}{lllll|llll|llll}
\hline
\multirow{2}{*}{Method} & \multicolumn{4}{l}{\qquad \qquad CVUSA\_val}     & \multicolumn{4}{l}{\qquad \qquad CVACT\_val} & \multicolumn{4}{l}{\qquad \qquad CVACT\_test} \\ \cline{2-13} 
                        & R@1   & R@5   & R@10  & R@1\% & R@1    & R@5   & R@10  & R@1\% & R@1   & R@5    & R@10   & R@1\% \\ \hline \hline
CVM-Net~\cite{hu2018CVPR} & 22.47 & 49.98 & 63.18 & 93.62 & 20.15  & 45& 56.87 & 87.57 &    4.06   &   16.89     &     24.66   &  56.38     \\ \hline
Liu, et al.~\cite{liu2019CVPR} & 40.79 & 66.82 & 76.36 & 96.12 & 46.96  & 68.28 & 75.48 & 92.01 & 19.9  & 34.82  & 41.23  & 63.79 \\ \hline
Regmi, et al.~\cite{regmi2019ICCV} & 48.75 & -  & 81.27 & 95.98 & -      & - & - & - & - & - & -& -    \\ \hline
CVFT~\cite{shi2020AAAI} & 61.43 & 84.69 & 90.49 & 99.02 & 61.05  & 81.33 & 86.52 & 95.93 & 34.39& 58.83& 66.78& 95.99\\ \hline
SAFA~\cite{shi2019NeurIPS}& 89.84 & 96.93 & 98.14 & 99.64 & 81.03  & 92.8  & 94.84 & 98.17 & 55.5  & 79.94  & 85.08  & 94.49 \\ \hline
DSM~\cite{shi2020CVPR} & 91.96 & 97.50  & \textbf{98.54} & \textbf{99.67} & 82.49  & 92.44 & 93.99 & 97.32 & 35.55 & 60.17 & 67.95 & 86.71  \\ \hline \hline
Ours    &   \textbf{92.56}   &    \textbf{97.55}   &  98.33     &   99.57    &   \textbf{83.28}     &  \textbf{93.57}     &   \textbf{95.42}    &   \textbf{98.22}    &   \textbf{61.29}    &  \textbf{85.13}      &     \textbf{89.14}   &  \textbf{98.32}    \\ \hline
\end{tabular}
}
\end{center}
\caption{A summary of our quantitative geo-localization experiments. We compare the recall-k (R@k) retrieval accuracy of our method with the current state-of-the-art on the CVUSA~\cite{zhai2017CVPR} and CVACT~\cite{liu2019CVPR} benchmarks.}
    \label{table:recallaccuracy}
\end{table*}
In our experiments, we evaluate the performance of our method both in terms of geo-localization and cross-view image synthesis. Overall, our results indicate that these two tasks reinforce each other and the joint training improves performance significantly. 
We present our quantitative results on cross-view geo-localization in~\Cref{subsec:retrieval_baselines} and on street view synthesis in~\Cref{subsec:synthesis_baselines}, with comparisons to state-of-the-art baselines. Furthermore, in~\Cref{subsec:synthesis_baselines}, we present qualitative results and comparisons. Finally, we provide an ablation study for further insights into how the different components of our method contribute to our results, see~\Cref{subsec:ablation_study}.
\subsection{Datasets} 
We consider the standard large scale cross-view benchmarks CVUSA~\cite{zhai2017CVPR} and CVACT~\cite{liu2019CVPR}, which consist of 44,416 and 137,218 pairs of top view satellite and panoramic street view images, respectively. Images depict streets of both rural and urban scenes. 
The orientation of the images is normalized, such that the north direction corresponds to the top part of satellite images and the center of the street-level images. 

For CVUSA, the first 35,532 ground-to-aerial image pairs are used for training and the remaining 8,884 pairs for validation. Additionally, the images in CVUSA are endowed with street view semantic segmentation labels, which we do not use since our method does not depend on any additional information. 

For the sake of consistency, the authors of CVACT~\cite{liu2019CVPR} chose the same training and validation set sizes as in CVUSA. The remaining 92,802 pairs comprise the test set. 
Additionally, the CVACT dataset provides UTM coordinates for each satellite-street pair. This large test set allows for a thorough investigation of the generalization ability of our proposed algorithm.

Moreover, since a lot of image pairs in CVACT were taken in close proximity to each other, for the test set of this benchmark, a retrieved satellite image is considered correct, as long as the distance to the actual ground truth match is less than 5 meters -- \ie, there might be multiple correct satellite matches for a given street level image.
%
%
\subsection{Cross-view geo-localization}\label{subsec:retrieval_baselines}
\paragraph{Recall metric}
For our geo-localization results on the CVUSA and CVACT benchmarks, we followed the standard evaluation protocol from prior work~\cite{liu2019CVPR,shi2019NeurIPS,shi2020CVPR,shi2020AAAI}. The main performance indicator here is the recall-k (R@k) retrieval accuracy.
Since our algorithm produces $L_2$ distances for each potential street view to satellite match, we can output a set of plausible matches. The R@k value is therefore defined as the ranking of the satellite ground-truth images which are correctly classified in one of the top-k matches for a given street view image. In particular, the R@1 metric measures the fraction of correct one-to-one matches.
\paragraph{Discussion}
~\Cref{table:recallaccuracy} shows our recall-k retrieval accuracies R@1, R@5, R@10, and R@1\%, in comparison to existing methods. There are two things we would like to point out: First of all, our method outperforms prior works on R@1 and R@5. The R@1 metric is a crucial criterion since it quantifies the percentage of exact matches. More importantly, our network shows strong results on the CVACT test set, which contains a large, city-scale collection of images. Specifically, our approach outperforms the best baselines by a significant margin for all considered metrics.
%

\subsection{Ground view image synthesis}\label{subsec:synthesis_baselines}
\begin{table*}[]
\begin{center}
\begin{tabular}{llllllll}
\hline
Method  & SSIM($\uparrow$)  & PSNR($\uparrow$)    & SD($\uparrow$)      & CS-1($\uparrow$) & CS-5($\uparrow$) & KL Scores($\downarrow$) & LPIPS($\downarrow$)  \\ \hline
Zhai, et al.~\cite{zhai2017CVPR}    & 0.414 & 11.502 & 10.631 & 13.97        & 42.09   & 27.43 \textpm\ 1.63   & -       \\ \hline
Pix2pix~\cite{isola2017CVPR} & 0.392 & 11.671 & 12.537 & 7.33           & 25.81         & 59.81 \textpm\ 2.12    & 0.595 \\ \hline
X-Seq~\cite{regmi2018CVPR}   & 0.423 & 12.820 & 12.451 & 15.98        & 42.91         & 15.52 \textpm\ 1.73  & 0.590 \\ \hline
X-Fork~\cite{regmi2018CVPR}  & 0.435 & 13.064 & 12.684 & 20.58         & 50.51           & 11.71 \textpm\ 1.55   & 0.609 \\ \hline
Ours    & \textbf{0.447} & \textbf{13.895} & \textbf{15.221} &  \textbf{33.23} & \textbf{65.85}&    \textbf{3.59 \textpm\ 0.92} & \textbf{0.474}\\ \hline
\end{tabular}
\end{center}
\caption{Quantitative experiments on image synthesis on CVUSA benchmark. $\uparrow$ indicates higher is better, vice versa $\downarrow$ indicates lower is better.}
    \label{table:synthesisaccuracy}
\end{table*}
\paragraph{Metrics} 
For street view synthesis, we use the PSNR, SSIM, and Sharpness difference (SD) metrics. These quantify the pixel level difference between synthesized and ground-truth street views in terms of primitive geometric properties, see~\cite{regmi2018CVPR,mathieu2015deep,wang2004image} for definitions. 
Additionally, we examine two task-specific metrics which were proposed by~\cite{regmi2018CVPR}. The idea is to assess the similarity of our generated images and the real street views by comparing the predictions of a separate image classifier. Specifically, the class predictions of the images are assigned to one of 365 different categories from the Places dataset~\cite{zhou2017TPAMI} with a standard AlexNet~\cite{krizhevsky2012NeurIPS} image classifier. 
We can then measure the top-1 and top-5 classification scores (CS-1 and CS-5) of the fake images. Additionally, we compute the KL-divergence between the class label distributions of the real and synthesized street images.
Finally, we compare the perceptual similarity score (LPIPS) by using the AlexNet~\cite{krizhevsky2012NeurIPS} backbone, see~\cite{zhang2018CVPR} for a definition.
\paragraph{Discussion} 
~\Cref{table:synthesisaccuracy} contains a summary of our image synthesis results in terms of the quality metrics mentioned above. We compare our generated images to the current state-of-the-art in cross-view image synthesis. 
The first baseline method~\cite{zhai2017CVPR} predicts auxiliary semantic segmentation labels on the satellite images, maps them to the street view, and uses them to generate a corresponding ground-level image.~\cite{regmi2018CVPR} introduced two different architectures: X-Fork and X-Seq.  Moreover,~\cite{regmi2018CVPR} shows comparisons to the generic image-to-image translation architecture Pix2pix~\cite{isola2017CVPR} on cross-view synthesis. Along the same lines, we provide comparisons to~\cite{isola2017CVPR} analogously to the setting discussed in~\cite{regmi2018CVPR}.

We want to point out that all the baselines we consider here require semantic segmentation masks during training. Remarkably, our method still outperforms these existing methods, even though it does not rely on supervision in terms of these semantic labels. 
%
\begin{figure}
\centering
    \includegraphics[width=\linewidth]{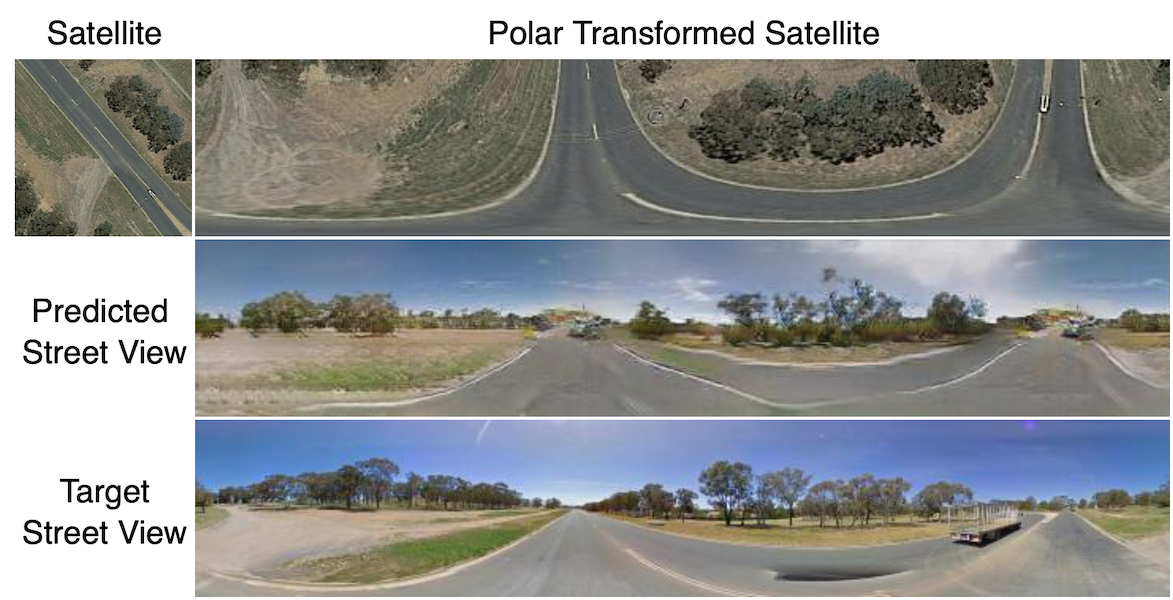}
    \caption{Qualitative result on CVACT. We show the input satellite-to-street pair, as well as the polar transformed satellite image and our synthesized street view. Note, that the considered baselines~\cite{regmi2018CVPR} cannot be applied here since there are no ground-truth annotated semantic maps on CVACT.}
    \label{fig:qualitative_cvact_comparison}
\end{figure}
\paragraph{Qualitative Experiments}\label{subsec:qualitative}
Our quantitative results show that our method indeed synthesizes more realistic and accurate street views than prior approaches. 
First, we present a qualitative comparison on the validation set of CVUSA in~\Cref{fig:qualitative_cvusa_comparison}. Note, that here we consider the current state-of-the-art baselines X-Seq and X-Fork which specialize on generating the first quarter of the street view. Additionally, we show qualitative results on the test set of CVACT in~\Cref{fig:qualitative_cvact_comparison}. On this benchmark, the other baselines unfortunately cannot be applied, since they require semantic segmentation labels during training which are not available for CVACT. 
Our method is able to generate highly plausible street-views, despite the fact that it uses less supervision than prior approaches~\cite{zhai2017CVPR, regmi2018CVPR}.
\subsection{Ablation Study}\label{subsec:ablation_study}
%
\paragraph{Geo-localization}
First of all, we investigate how the different components of our method affect the retrieval performance on CVUSA~\cite{zhai2017CVPR}. To that end, we perform the following ablations and report the results in~\Cref{table:ablationrecall}: 
The central question here is how much the image synthesis branch impacts the geo-localization task. 
First, we train the retrieval branch of our network without the generator \emph{decoder head} and the \emph{discriminator} network (\textbf{i.}). This means that the local features extracted by the generator encoder are only passed on to the $R$ branch and no street-view images are generated. This modified network leads to a lower recall accuracy since it cannot use the reinforced features from the image synthesis. The second experiment again omits the discriminator network and the GAN loss $\mathcal{L}_{cGAN}$ but still predicts a street view, solely based on the $\mathcal{L}_1$ reconstruction loss (\textbf{ii.}). This modification leads to a less significant drop in accuracy, since the reconstruction of the street view supports the retrieval part at least to some extent.
Another aspect we want to examine (\textbf{iii.}) is how the retrieval performance alters if we pass on the generated images $G(I_{\mathrm{ps}})$ to the retrieval branch $R$ instead of the latent bottleneck features from $G_{E}$. Here, we again observe a decrease in performance. The generated images themselves are simply not sufficient to convey the same information richness as the bottleneck features -- ultimately, the retrieval branch has less latent information available.
Overall, the results in~\Cref{table:ablationrecall} suggest that the image synthesis indeed benefits the retrieval accuracy. This can by and large be attributed to the fact that learning to generate cross-view images yields local features that are more coherent across the different input domains.
%
%
\begin{table}[H]
\begin{center}
\begin{tabular}{lllll}
\hline & \multicolumn{4}{l}{\qquad \qquad \quad CVUSA} \\ \cline{2-5} 
\multirow{-2}{*}{Method} & R@1 & R@5 & R@10 & R@1\%      \\ \hline
\\[-1em]
\textbf{i.} w/o $G\&D$ & 88.06& 96.47& 97.88 & 99.62 \\ \hline
\\[-1em]
\textbf{ii.} w/o $\mathcal{L}_{cGAN}$ &   91.92& 97.22& 98.29& 99.65  \\ \hline
\\[-1em]
\textbf{iii.} w/ $G(I_{ps})$ & 89.98&96.78& 98.04& \textbf{99.66}\\ \hline
\\[-1em]
Ours                     &\textbf{92.56} & \textbf{97.55} & \textbf{98.33} & 99.57 \\ \hline
\\[-1em]
\end{tabular}
\end{center}
\caption{Ablation study on our geo-localization experiments. We show the retrieval recall-k (R@k) accuracies for different versions of our full pipeline, see~\Cref{subsec:ablation_study} for more details.} 
    \label{table:ablationrecall}
\end{table}
\paragraph{Street view synthesis}
For the image synthesis task, we consider the following ablations of our full pipeline, see~\Cref{table:ablationsynthesis} for a summary of the results: First of all, we measure the image quality of the pure, polar-transformed satellite images in comparison to the input street images. These results confirm that a simple geometric coordinate transformation is clearly inferior to a learned, generative model. Furthermore, we train our image generation branch without the retrieval head $R$.  
These results suggest that joint training indeed benefits the image synthesis task.
The reason for that is, again, that the multi-task learning incentivizes our network to learn superior features which ultimately improves the performance of both tasks at the same time.
\begin{table}[H]
\begin{center}
\begin{tabular}{llll}
\hline & \multicolumn{3}{l}{\quad \quad CVUSA} \\ \cline{2-4} 
\multirow{-2}{*}{Method} & SSIM$\uparrow$ & PSNR$\uparrow$ & SD$\uparrow$    \\ \hline
\\[-1em]
$I_{\mathrm{ps}}$ vs $I_{\mathrm{g}}$ & 0.2892   & 10.7325 & 14.2291\\\hline
\\[-1em]
w/o $R$                  &0.4392 & 13.6858 & 15.0843 \\ \hline
\\[-1em]
Ours                     &\textbf{0.4472} & \textbf{13.8952} &   \textbf{15.2215}\\ \hline
\\[-1em]
\end{tabular}
\end{center}
\caption{Ablation study on our image synthesis experiments. This shows, that our generator produces the most accurate images, in combination with the retrieval branch.}
    \label{table:ablationsynthesis}
\end{table}
\section{Conclusion}
We presented ``Coming Down to Earth'', a new framework for cross-view image-based geo-localization. Our model integrates image synthesis and retrieval in one architecture which is end-to-end trainable.
The key insight is that satellite-to-street view synthesis promotes a latent feature space that is coherent across the two input domains, which benefits the localization.
The image retrieval branch, on the other hand, naturally incentivizes the generator to create images that faithfully depict the content of the scene.
Remarkably, our method outperforms existing cross-view synthesis approaches, even though it does not rely on any additional semantic information. Finally, we obtain state-of-the-art performance in terms of cross-view geo-localization on both considered benchmarks CVUSA and CVACT.
\vspace{1.5em}
\paragraph{Acknowledgements}
We would like to thank Marvin Eisenberger for valuable discussions. This research was supported by the Humboldt Foundation through the Sofja Kovalevskaja Award and the Helmholtz Association under the joint research school ``Munich School for Data Science - MUDS''.
\clearpage
\appendix
\section{Architecture details}
\begin{figure}
\centering
    \includegraphics[scale=0.19]{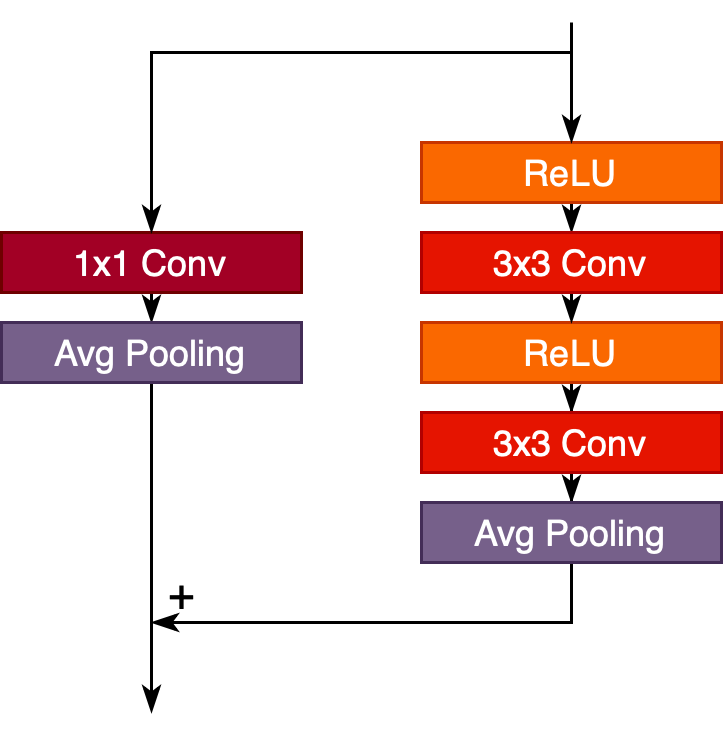}
    \caption{Residual downsampling blocks.}
    \label{fig:residual_block_down}
\end{figure}

\begin{figure}
\centering
    \includegraphics[scale=0.19]{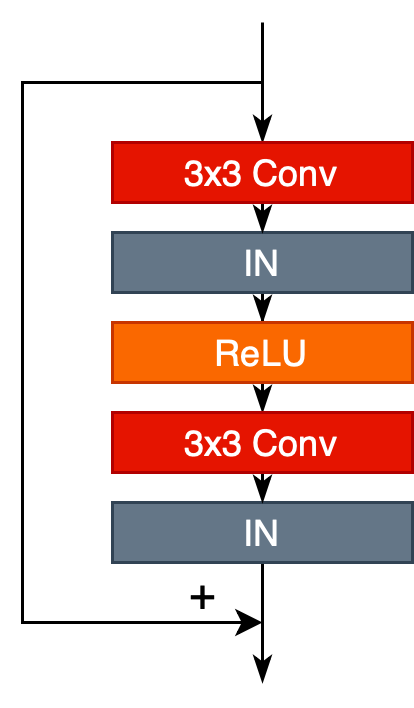}
    \caption{Residual blocks.}
    \label{fig:residual_block}
\end{figure}

\begin{figure}
\centering
    \includegraphics[scale=0.19]{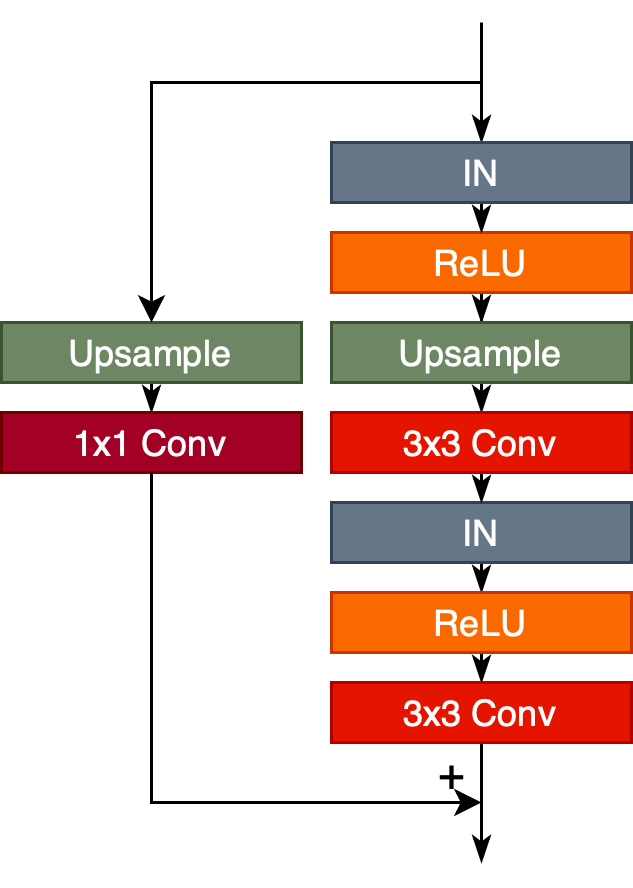}
    \caption{Residual upsampling blocks.}
    \label{fig:residual_block_up}
\end{figure}

We outlined the general structure of our model in the main paper.
Here, we provide more technical details about our network in terms of the generator $G$, discriminator $D$, and the retrieval $R$ branches.
\paragraph{Generator.}
The generator $G$ is constructed as a U-Net~\cite{ronneberger2015unet} architecture consisting of residual blocks~\cite{he2016CVPR}. This design is inspired by existing image-to-image translation models~\cite{brock2018large, zhang2019ICML, maximov2020CVPR} with similar architectures. We provide the exact components of our generator in~\Cref{table:generator_detail}. For a given satellite image, $G$ uses residual downsampling blocks (see~\Cref{fig:residual_block_down}) that in total reduce the spatial size by a factor of 64. On the reduced resolution, our bottleneck refines the features with 6 residual blocks, see~\Cref{fig:residual_block}. The last part of $G$ then consists of residual upsampling blocks (see~\Cref{fig:residual_block_up}) that mirror the downsampling blocks from before, such that the inputs and outputs of $G$ have the same spatial size. Additionally, we add a self-attention (non-local) block~\cite{wang2018non} after the first upsampling block. This strategy helps to learn global dependencies in the image, as was shown in prior work~\cite{zhang2019ICML, brock2018large}. Finally, we use an instance normalization layer (IN)~\cite{ulyanov2016instance} after each upsampling and downsampling block and spectral normalization~\cite{miyato2018spectral} after each convolution layer.
\paragraph{Discriminator.}
The discriminator $D$ is designed as a modified PatchGAN classifier~\cite{isola2017CVPR}, see~\Cref{table:discriminator_detail}. In this context, the satellite-street input pair comprises the {\it real} input whereas the {\it fake} input consists of the input satellite and generated street $G(I_{ps})$ images. Similarly to the generator, we use non-local self-attention blocks on the 28x154 resolution. We also again use spectral normalization~\cite{miyato2018spectral} after each convolution layer, which regularizes each individual set of features to a spectral radius of 1. Note, that spectral normalization is not used after the last convolution layer of both the generator and discriminator. 
\paragraph{Retrieval.}
The retrieval branch $R$ of our network determines the corresponding satellite image for a given street panorama. This is done by finding a global feature encoding for both satellite and street view input images. The local features for the satellite inputs are not computed here, because we can reuse the features from the encoder part of the generator $G_{E}(I_{ps})$. Specifically, we use the output from \emph{the last residual block in the bottleneck}.
In order to obtain an equivalent set of features for the street-view input images, we use a modified ResNet34~\cite{he2016CVPR} feature extractor. 
This yields a set of features for both inputs with the same spatial and channel sizes. Afterward, we convert the local features of both inputs to global descriptors by using the spatial attention module $SA$ which we explain in the main paper.
\begin{table}[h!]
\begin{center}
\begin{tabular}{l}
\hline
\qquad \qquad \quad \emph{Generator}\\ \hline \hline
Satellite (3, 112, 616)                                                                 \\ \hline
\\[-1em]
Conv (32, 112, 616) + IN                                                               \\ \hline
\\[-1em]
\emph{(enc1)} Resblock Down (64, 56, 308) + IN                                                   \\ \hline
\\[-1em]
\emph{(enc2)} Resblock Down (128, 28, 154) + IN                                                  \\ \hline
\\[-1em]
\emph{(enc3)} Resblock Down (256, 14, 77) + IN                                                   \\ \hline
\\[-1em]
Resblock (256, 14, 77) x 6                                                              \\ \hline
\\[-1em]
\begin{tabular}[c]{@{}l@{}}+ \emph{concat (enc3)}\\ Resblock Up (128, 28, 154) + IN\end{tabular} \\ \hline
\\[-1em]
\begin{tabular}[c]{@{}l@{}}+ \emph{concat (enc2)}\\ Non-local Block (256, 28, 154 )\end{tabular} \\ \hline
\\[-1em]
Resblock Up (64, 56, 308) + IN                                                          \\ \hline
\\[-1em]
\begin{tabular}[c]{@{}l@{}}+ \emph{concat (enc1)}\\ Resblock Up (64, 112, 616) + IN\end{tabular} \\ \hline
\\[-1em]
Conv (3, 112, 616) + Tanh                                                            \\ \hline
\\[-1em]
\end{tabular}
\end{center}
\caption{Exact technical specifications of our generator $G$.}
    \label{table:generator_detail}
\end{table}
\begin{table}[]
\begin{center}
\begin{tabular}{l}
\hline
\qquad \qquad \quad \emph{Discriminator}                            \\ \hline \hline
\\[-1em]
Satellite+Street (6, 112, 616)           \\ \hline
\\[-1em]
4x4 Conv + LeakyReLU(0.1) (64, 56, 308)  \\ \hline
\\[-1em]
4x4 Conv + LeakyReLU(0.1) (128, 28, 154) \\ \hline
\\[-1em]
Non-local Block (128, 28, 154)           \\ \hline
\\[-1em]
4x4 Conv + LeakyReLU(0.1) (256, 14, 77)  \\ \hline
\\[-1em]
4x4 Conv + LeakyReLU(0.1) (512, 14, 76)  \\ \hline
\\[-1em]
4x4 Conv (1, 14, 75)   \\ \hline
\\[-1em]
\end{tabular}
\end{center}
\caption{Exact structure of our discriminator module $D$.}
    \label{table:discriminator_detail}
\end{table}
\section{Implementation details}
We implement our network in PyTorch using Adam optimizer~\cite{kingma2014adam}. The momentum parameters $\beta1$ and $\beta2$ are set to 0.5 and 0.999, respectively, and the learning rate for all three networks is set to $1e-4$. 
The resolution of both the ground images and polar transformed satellite images is 112x616. We augment both the street view and the polar transformed satellite images with random horizontal flips.
Furthermore, we normalize the pixel intensity values to the interval [-1, 1]. 
For the weighted soft-margin loss, we use the exhaustive mini-batch strategy to create the triplets within a batch. For the batch size $B$ (we choose $B=32$), this strategy sums up the triplet loss for all $2B(B-1)$ combinations of positive and negative pairs, see~\cite{vo2016ECCV} for more details. Moreover, we use a hard negative mining strategy after the loss converged as a fine-tuning step during training. Specifically, we sort all triplets in the current batch by relevance, i.e., the loss value, and discard a certain percentage of the triplets with the least amount of surplus information. 
For the spatial aware feature aggregation, we use $k=8$ attention masks.
Finally, we choose the hyperparameter from our weighted soft margin loss as $\alpha=10$.
During training, we follow the standard protocol for GAN optimization~\cite{goodfellow2014NeurIPS}. 
In particular, we alternate between updating the weights of each network individually with one update step per cycle.
The gradients for the generator are coming from both the discriminator and retrieval networks.
Additionally, the weights for the individual loss functions are set to: $\lambda_{ret} = 1000$, $\lambda_{L_1} = 100$ and $\lambda_{GAN} = 1$. 
\begin{figure*}
    \centering
    \includegraphics[width=\linewidth]{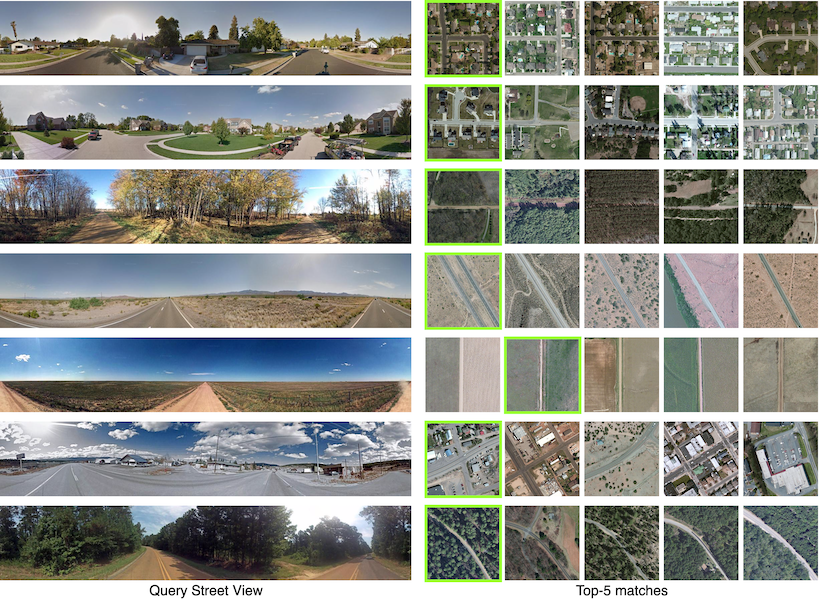}
    \caption{Geo-localization results on CVUSA~\cite{zhai2017CVPR}. For a given query street view (left), we show the closest satellite matches produced by our method. Green boxes denote the ground truth match.}
    \label{fig:cvusa_recall}
\end{figure*}
\begin{figure*}
\centering
    \includegraphics[width=\linewidth]{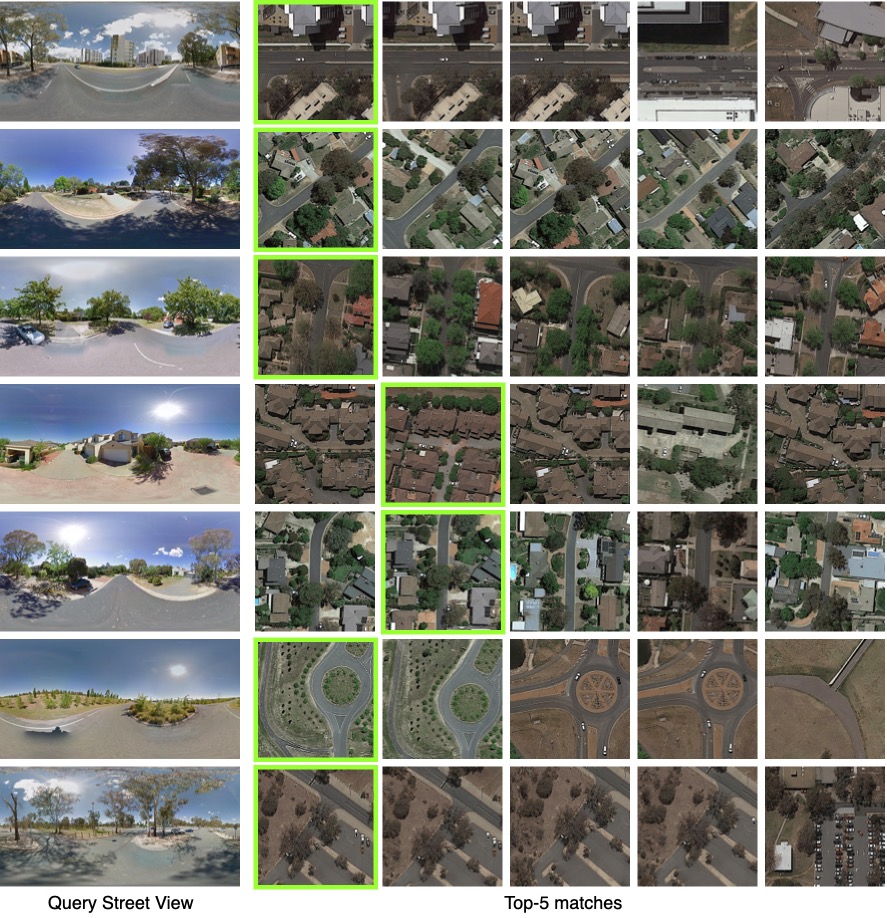}
    \caption{Geo-localization results on the CVACT~\cite{liu2019CVPR} test set. For each street image, we show the closest five matches predicted by our method, as well as the ground truth match (green box). Since the coverage of the CVACT dataset is quite dense, there are at times multiple matches that are considered correct, as long as the distance to the ground truth match is less than 5 meters (\eg, in the last row, the second satellite image is also a correct match). Our method consistently retrieves not only the closest match, but also multiple images in the same region which yields a robust localization performance.}
    \label{fig:cvact_test_recall}
\end{figure*}
\begin{figure*}
\centering
    \includegraphics[width=\linewidth]{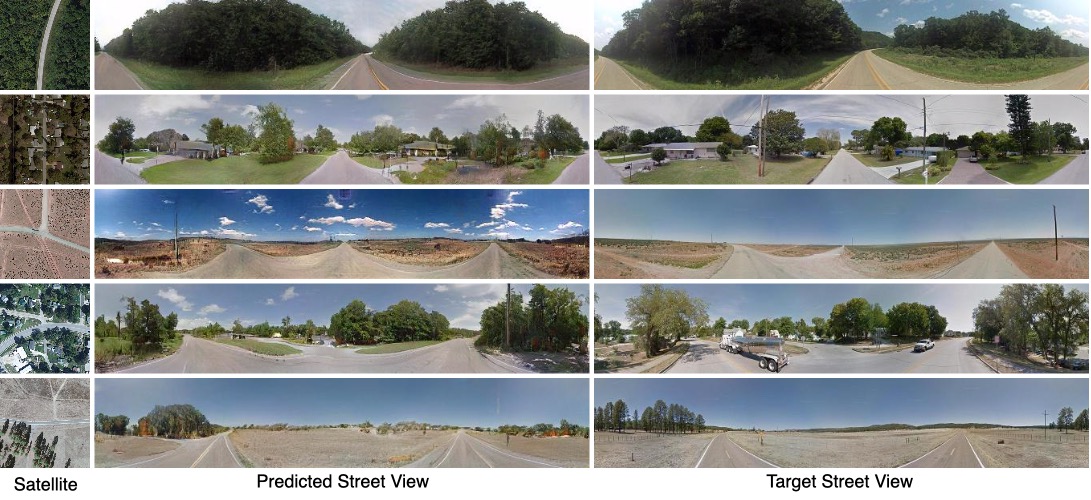}
    \caption{Qualitative cross-view synthesis examples on the CVUSA dataset.}
    \label{fig:cvusa_synthesis}
\end{figure*}
\begin{figure*}
\centering
    \includegraphics[width=\linewidth]{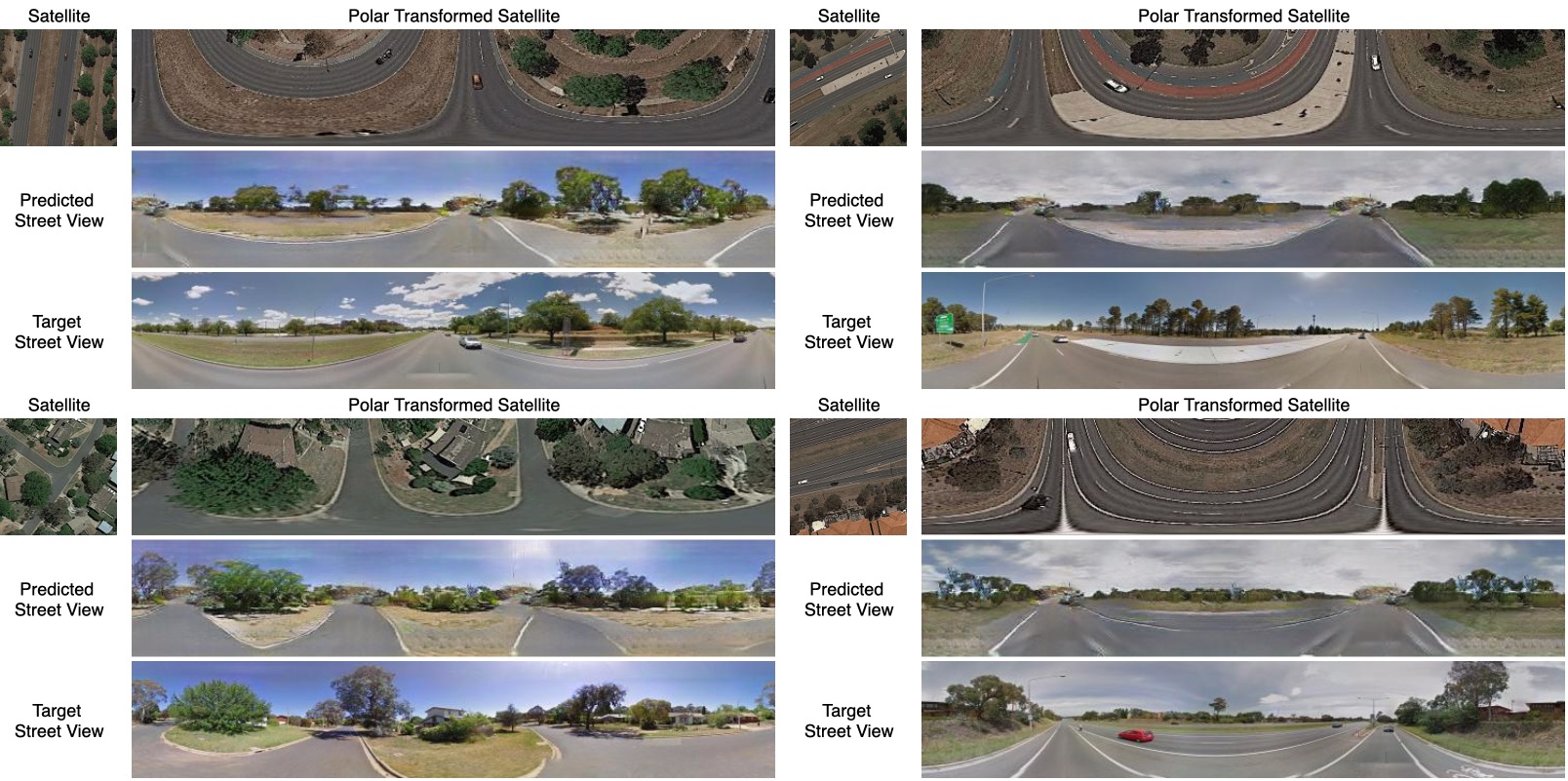}
    \caption{Qualitative examples on the CVACT dataset. The two examples on the left side are from the validation set and the ones on the right side from the large-scale test set.}
    \label{fig:cvact_synthesis}
\end{figure*}
\begin{figure*}[h]
\centering
    \includegraphics[width=\linewidth]{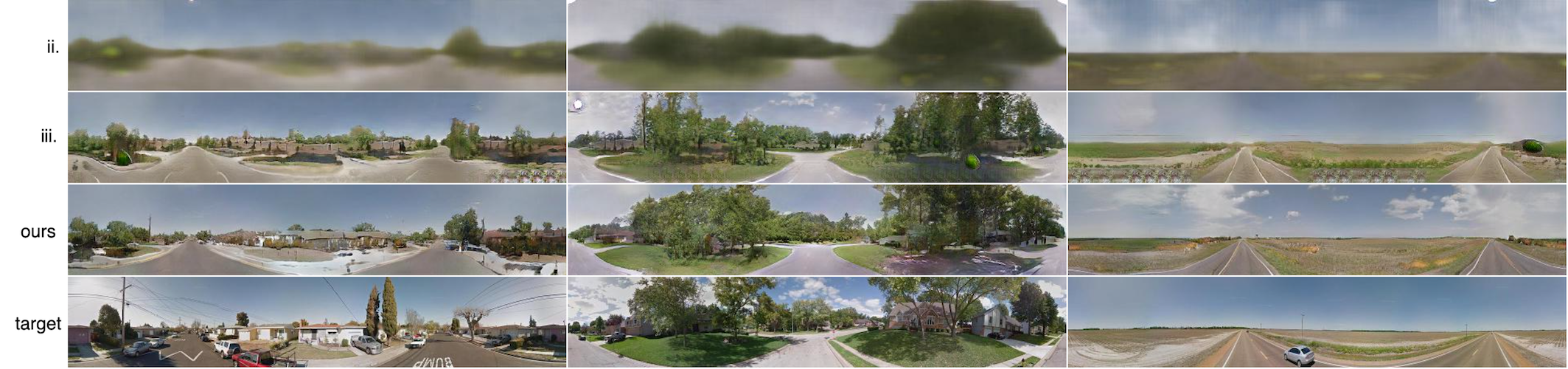}
    \caption{Qualitative comparisons corresponding to the results from our ablation study  in~\Cref{table:ablationrecall}. We show the synthesized images for three different examples for ablations \textbf{ii.} and \textbf{iii.}, our method and the ground truth street views.}
    \label{fig:ablation_qualitative}
\end{figure*}
\section{Additional qualitative results}
\paragraph{Geo-localization.}
In the localization branch $R$, our algorithm produces $L_2$ distances between the features of satellite-street pairs. This means that for a given street image, our algorithm can output a set of the closest satellite matches and rank them according to plausibility. For a given street image, the recall-$k$ retrieval accuracy (R@$k$) then measures whether the ground-truth satellite pair is among the first $k$ predicted matches.
Here, we visualize some of the closest satellite images for a given query street image for examples from CVUSA (see~\Cref{fig:cvusa_recall}) and the CVACT test set (see~\Cref{fig:cvact_test_recall}).
\paragraph{Cross-view synthesis.}
For a more complete picture, we show additional qualitative results for satellite-to-street view synthesis for both considered benchmarks in~\Cref{fig:cvusa_synthesis} and~\Cref{fig:cvact_synthesis}.
\paragraph{Ablation Study.}
We now assess the qualitative difference between our full architecture and the ablations \textbf{ii.} and \textbf{iii.} in~\Cref{table:ablationrecall}, see~\Cref{fig:ablation_qualitative}. Even though the ablation \textbf{ii.} only slightly underperforms our main pipeline in terms of the image retrieval task, the synthesized images are significantly less realistic. The $\mathcal{L}_1$ loss itself is sufficient to represent low frequencies, which yields latent features that represent the overall image structure, but lacks high frequency details producing blurry images. The qualitative comparison between ablation \textbf{iii.} and our main pipeline illustrates that $\mathcal{L}_{cGAN}$ synthesizes photo-realistic street-views. On the other hand, the qualitative difference between them highlights the importance of using the latent representation in our full architecture.
\clearpage
\clearpage
\clearpage
{\small
\bibliographystyle{ms}
\bibliography{ms}

\begin{thebibliography}{10}\itemsep=-1pt

\bibitem{arandjelovic2016CVPR}
Relja Arandjelovic, Petr Gronat, Akihiko Torii, Tomas Pajdla, and Josef Sivic.
\newblock Netvlad: Cnn architecture for weakly supervised place recognition.
\newblock In {\em Proceedings of the IEEE conference on computer vision and
  pattern recognition}, pages 5297--5307, 2016.

\bibitem{arandjelovic2014ACCV}
Relja Arandjelovi{\'c} and Andrew Zisserman.
\newblock Dislocation: Scalable descriptor distinctiveness for location
  recognition.
\newblock In {\em Asian Conference on Computer Vision}, pages 188--204.
  Springer, 2014.

\bibitem{bansal2011geo}
Mayank Bansal, Harpreet~S Sawhney, Hui Cheng, and Kostas Daniilidis.
\newblock Geo-localization of street views with aerial image databases.
\newblock In {\em Proceedings of the 19th ACM international conference on
  Multimedia}, pages 1125--1128, 2011.

\bibitem{brock2018large}
Andrew Brock, Jeff Donahue, and Karen Simonyan.
\newblock Large scale gan training for high fidelity natural image synthesis.
\newblock {\em arXiv preprint arXiv:1809.11096}, 2018.

\bibitem{cai2019ICCV}
Sudong Cai, Yulan Guo, Salman Khan, Jiwei Hu, and Gongjian Wen.
\newblock Ground-to-aerial image geo-localization with a hard exemplar
  reweighting triplet loss.
\newblock In {\em Proceedings of the IEEE International Conference on Computer
  Vision}, pages 8391--8400, 2019.

\bibitem{castaldo2015CVPRW}
Francesco Castaldo, Amir Zamir, Roland Angst, Francesco Palmieri, and Silvio
  Savarese.
\newblock Semantic cross-view matching.
\newblock In {\em Proceedings of the IEEE International Conference on Computer
  Vision Workshops}, pages 9--17, 2015.

\bibitem{chen2011CVPR}
David~M Chen, Georges Baatz, Kevin K{\"o}ser, Sam~S Tsai, Ramakrishna
  Vedantham, Timo Pylv{\"a}n{\"a}inen, Kimmo Roimela, Xin Chen, Jeff Bach, Marc
  Pollefeys, et~al.
\newblock City-scale landmark identification on mobile devices.
\newblock In {\em CVPR 2011}, pages 737--744. IEEE, 2011.

\bibitem{deng2009CVPR}
Jia Deng, Wei Dong, Richard Socher, Li-Jia Li, Kai Li, and Li Fei-Fei.
\newblock Imagenet: A large-scale hierarchical image database.
\newblock In {\em 2009 IEEE conference on computer vision and pattern
  recognition}, pages 248--255. Ieee, 2009.

\bibitem{goodfellow2014NeurIPS}
Ian Goodfellow, Jean Pouget-Abadie, Mehdi Mirza, Bing Xu, David Warde-Farley,
  Sherjil Ozair, Aaron Courville, and Yoshua Bengio.
\newblock Generative adversarial nets.
\newblock In {\em Advances in neural information processing systems}, pages
  2672--2680, 2014.

\bibitem{hays2008CVPR}
James Hays and Alexei~A Efros.
\newblock Im2gps: estimating geographic information from a single image.
\newblock In {\em 2008 IEEE conference on computer vision and pattern
  recognition}, pages 1--8. IEEE, 2008.

\bibitem{he2016CVPR}
Kaiming He, Xiangyu Zhang, Shaoqing Ren, and Jian Sun.
\newblock Deep residual learning for image recognition.
\newblock In {\em Proceedings of the IEEE conference on computer vision and
  pattern recognition}, pages 770--778, 2016.

\bibitem{hu2018CVPR}
Sixing Hu, Mengdan Feng, Rang~MH Nguyen, and Gim Hee~Lee.
\newblock Cvm-net: Cross-view matching network for image-based ground-to-aerial
  geo-localization.
\newblock In {\em Proceedings of the IEEE Conference on Computer Vision and
  Pattern Recognition}, pages 7258--7267, 2018.

\bibitem{isola2017CVPR}
Phillip Isola, Jun-Yan Zhu, Tinghui Zhou, and Alexei~A Efros.
\newblock Image-to-image translation with conditional adversarial networks.
\newblock In {\em Proceedings of the IEEE conference on computer vision and
  pattern recognition}, pages 1125--1134, 2017.

\bibitem{kingma2014adam}
Diederik~P Kingma and Jimmy Ba.
\newblock Adam: A method for stochastic optimization.
\newblock {\em arXiv preprint arXiv:1412.6980}, 2014.

\bibitem{krizhevsky2012NeurIPS}
Alex Krizhevsky, Ilya Sutskever, and Geoffrey~E Hinton.
\newblock Imagenet classification with deep convolutional neural networks.
\newblock In {\em Advances in neural information processing systems}, pages
  1097--1105, 2012.

\bibitem{li2016ECCV}
Chuan Li and Michael Wand.
\newblock Precomputed real-time texture synthesis with markovian generative
  adversarial networks.
\newblock In {\em European conference on computer vision}, pages 702--716.
  Springer, 2016.

\bibitem{lin2013CVPR}
Tsung-Yi Lin, Serge Belongie, and James Hays.
\newblock Cross-view image geolocalization.
\newblock In {\em Proceedings of the IEEE Conference on Computer Vision and
  Pattern Recognition}, pages 891--898, 2013.

\bibitem{lin2015CVPR}
Tsung-Yi Lin, Yin Cui, Serge Belongie, and James Hays.
\newblock Learning deep representations for ground-to-aerial geolocalization.
\newblock In {\em Proceedings of the IEEE conference on computer vision and
  pattern recognition}, pages 5007--5015, 2015.

\bibitem{liu2019CVPR}
Liu Liu and Hongdong Li.
\newblock Lending orientation to neural networks for cross-view
  geo-localization.
\newblock In {\em Proceedings of the IEEE conference on computer vision and
  pattern recognition}, pages 5624--5633, 2019.

\bibitem{lu2020CVPR}
Xiaohu Lu, Zuoyue Li, Zhaopeng Cui, Martin~R Oswald, Marc Pollefeys, and
  Rongjun Qin.
\newblock Geometry-aware satellite-to-ground image synthesis for urban areas.
\newblock In {\em Proceedings of the IEEE/CVF Conference on Computer Vision and
  Pattern Recognition}, pages 859--867, 2020.

\bibitem{mathieu2015deep}
Michael Mathieu, Camille Couprie, and Yann LeCun.
\newblock Deep multi-scale video prediction beyond mean square error.
\newblock {\em arXiv preprint arXiv:1511.05440}, 2015.

\bibitem{maximov2020CVPR}
Maxim Maximov, Ismail Elezi, and Laura Leal-Taix{\'e}.
\newblock Ciagan: Conditional identity anonymization generative adversarial
  networks.
\newblock In {\em Proceedings of the IEEE/CVF Conference on Computer Vision and
  Pattern Recognition}, pages 5447--5456, 2020.

\bibitem{miyato2018spectral}
Takeru Miyato, Toshiki Kataoka, Masanori Koyama, and Yuichi Yoshida.
\newblock Spectral normalization for generative adversarial networks.
\newblock {\em arXiv preprint arXiv:1802.05957}, 2018.

\bibitem{mousavian2016semantic}
Arsalan Mousavian and Jana Kosecka.
\newblock Semantic image based geolocation given a map.
\newblock {\em arXiv preprint arXiv:1609.00278}, 2016.

\bibitem{regmi2018CVPR}
Krishna Regmi and Ali Borji.
\newblock Cross-view image synthesis using conditional gans.
\newblock In {\em Proceedings of the IEEE Conference on Computer Vision and
  Pattern Recognition}, pages 3501--3510, 2018.

\bibitem{regmi2019CVIU}
Krishna Regmi and Ali Borji.
\newblock Cross-view image synthesis using geometry-guided conditional gans.
\newblock {\em Computer Vision and Image Understanding}, 187:102788, 2019.

\bibitem{regmi2019ICCV}
Krishna Regmi and Mubarak Shah.
\newblock Bridging the domain gap for ground-to-aerial image matching.
\newblock In {\em Proceedings of the IEEE International Conference on Computer
  Vision}, pages 470--479, 2019.

\bibitem{ronneberger2015unet}
Olaf Ronneberger, Philipp Fischer, and Thomas Brox.
\newblock U-net: Convolutional networks for biomedical image segmentation.
\newblock In {\em International Conference on Medical image computing and
  computer-assisted intervention}, pages 234--241. Springer, 2015.

\bibitem{sattler2016CVPR}
Torsten Sattler, Michal Havlena, Konrad Schindler, and Marc Pollefeys.
\newblock Large-scale location recognition and the geometric burstiness
  problem.
\newblock In {\em Proceedings of the IEEE Conference on Computer Vision and
  Pattern Recognition}, pages 1582--1590, 2016.

\bibitem{schindler2007CVPR}
Grant Schindler, Matthew Brown, and Richard Szeliski.
\newblock City-scale location recognition.
\newblock In {\em 2007 IEEE Conference on Computer Vision and Pattern
  Recognition}, pages 1--7. IEEE, 2007.

\bibitem{shi2019NeurIPS}
Yujiao Shi, Liu Liu, Xin Yu, and Hongdong Li.
\newblock Spatial-aware feature aggregation for image based cross-view
  geo-localization.
\newblock In {\em Advances in Neural Information Processing Systems}, pages
  10090--10100, 2019.

\bibitem{shi2020CVPR}
Yujiao Shi, Xin Yu, Dylan Campbell, and Hongdong Li.
\newblock Where am i looking at? joint location and orientation estimation by
  cross-view matching.
\newblock In {\em Proceedings of the IEEE/CVF Conference on Computer Vision and
  Pattern Recognition}, pages 4064--4072, 2020.

\bibitem{shi2020AAAI}
Yujiao Shi, Xin Yu, Liu Liu, Tong Zhang, and Hongdong Li.
\newblock Optimal feature transport for cross-view image geo-localization.
\newblock In {\em AAAI}, pages 11990--11997, 2020.

\bibitem{torii2015CVPR}
Akihiko Torii, Relja Arandjelovic, Josef Sivic, Masatoshi Okutomi, and Tomas
  Pajdla.
\newblock 24/7 place recognition by view synthesis.
\newblock In {\em Proceedings of the IEEE Conference on Computer Vision and
  Pattern Recognition}, pages 1808--1817, 2015.

\bibitem{ulyanov2016instance}
Dmitry Ulyanov, Andrea Vedaldi, and Victor Lempitsky.
\newblock Instance normalization: The missing ingredient for fast stylization.
\newblock {\em arXiv preprint arXiv:1607.08022}, 2016.

\bibitem{vo2016ECCV}
Nam~N Vo and James Hays.
\newblock Localizing and orienting street views using overhead imagery.
\newblock In {\em European conference on computer vision}, pages 494--509.
  Springer, 2016.

\bibitem{wang2018non}
Xiaolong Wang, Ross Girshick, Abhinav Gupta, and Kaiming He.
\newblock Non-local neural networks.
\newblock In {\em Proceedings of the IEEE conference on computer vision and
  pattern recognition}, pages 7794--7803, 2018.

\bibitem{wang2004image}
Zhou Wang, Alan~C Bovik, Hamid~R Sheikh, and Eero~P Simoncelli.
\newblock Image quality assessment: from error visibility to structural
  similarity.
\newblock {\em IEEE transactions on image processing}, 13(4):600--612, 2004.

\bibitem{workman2015CVPRW}
Scott Workman and Nathan Jacobs.
\newblock On the location dependence of convolutional neural network features.
\newblock In {\em Proceedings of the IEEE Conference on Computer Vision and
  Pattern Recognition Workshops}, pages 70--78, 2015.

\bibitem{workman2015ICCV}
Scott Workman, Richard Souvenir, and Nathan Jacobs.
\newblock Wide-area image geolocalization with aerial reference imagery.
\newblock In {\em Proceedings of the IEEE International Conference on Computer
  Vision}, pages 3961--3969, 2015.

\bibitem{zamir2010ECCV}
Amir~Roshan Zamir and Mubarak Shah.
\newblock Accurate image localization based on google maps street view.
\newblock In {\em European Conference on Computer Vision}, pages 255--268.
  Springer, 2010.

\bibitem{zamir2014image}
Amir~Roshan Zamir and Mubarak Shah.
\newblock Image geo-localization based on multiplenearest neighbor feature
  matching usinggeneralized graphs.
\newblock {\em IEEE transactions on pattern analysis and machine intelligence},
  36(8):1546--1558, 2014.

\bibitem{zhai2017CVPR}
Menghua Zhai, Zachary Bessinger, Scott Workman, and Nathan Jacobs.
\newblock Predicting ground-level scene layout from aerial imagery.
\newblock In {\em Proceedings of the IEEE Conference on Computer Vision and
  Pattern Recognition}, pages 867--875, 2017.

\bibitem{zhang2019ICML}
Han Zhang, Ian Goodfellow, Dimitris Metaxas, and Augustus Odena.
\newblock Self-attention generative adversarial networks.
\newblock In {\em International Conference on Machine Learning}, pages
  7354--7363. PMLR, 2019.

\bibitem{zhang2018CVPR}
Richard Zhang, Phillip Isola, Alexei~A Efros, Eli Shechtman, and Oliver Wang.
\newblock The unreasonable effectiveness of deep features as a perceptual
  metric.
\newblock In {\em Proceedings of the IEEE conference on computer vision and
  pattern recognition}, pages 586--595, 2018.

\bibitem{zhou2017TPAMI}
Bolei Zhou, Agata Lapedriza, Aditya Khosla, Aude Oliva, and Antonio Torralba.
\newblock Places: A 10 million image database for scene recognition.
\newblock {\em IEEE transactions on pattern analysis and machine intelligence},
  40(6):1452--1464, 2017.

\bibitem{zhou2014NeurIPS}
Bolei Zhou, Agata Lapedriza, Jianxiong Xiao, Antonio Torralba, and Aude Oliva.
\newblock Learning deep features for scene recognition using places database.
\newblock In {\em Advances in neural information processing systems}, pages
  487--495, 2014.

\bibitem{zhu20183DV}
Xinge Zhu, Zhichao Yin, Jianping Shi, Hongsheng Li, and Dahua Lin.
\newblock Generative adversarial frontal view to bird view synthesis.
\newblock In {\em International conference on 3D Vision (3DV)}, pages 454--463.
  IEEE, 2018.

\end{thebibliography}
}
\end{document}